\documentclass[letterpaper]{article} 
\usepackage{aaai25}  
\usepackage{times}  
\usepackage{helvet}  
\usepackage{courier}  
\usepackage[hyphens]{url}  
\usepackage{graphicx} 
\usepackage{natbib}  
\usepackage{caption} 
\usepackage{algorithm}
\usepackage{graphicx}
\usepackage{subcaption}
\usepackage{graphicx}

\usepackage{algpseudocode}
\usepackage{amsmath} 
\usepackage{amssymb}

\usepackage{graphicx}
\usepackage{inconsolata}
\usepackage{microtype}
\usepackage{array} 
\usepackage{amssymb} 
\usepackage{amsmath}
\usepackage{booktabs}
\usepackage{colortbl}
\usepackage[table]{xcolor}
\usepackage{makecell}
\usepackage{pifont}

\usepackage{tabularx}

\usepackage{newfloat}
\usepackage{listings}
\DeclareCaptionStyle{ruled}{labelfont=normalfont,labelsep=colon,strut=off} 
\floatstyle{ruled}
\newfloat{listing}{tb}{lst}{}
\floatname{listing}{Listing}
\title{Fine-grained and Explainable Factuality Evaluation \\ for Multimodal Summarization}
\author {
    Yue Zhang, Jingxuan Zuo, Ke Su, Liqiang Jing
}
\affiliations {
    The University of Texas at Dallas
}

\begin{document}

\maketitle

\begin{abstract}

Multimodal summarization aims to generate a concise summary based on the input text and image. However, the existing methods potentially suffer from unfactual output. To evaluate the factuality of multimodal summarization models, we propose two fine-grained and explainable evaluation frameworks (FALLACIOUS) for different application scenarios, i.e. reference-based factuality evaluation framework and reference-free factuality evaluation framework. Notably, the reference-free factuality evaluation framework doesn't need ground truth and hence it has a wider application scenario. To evaluate the effectiveness of the proposed frameworks, we compute the correlation between our frameworks and the other metrics. The experimental results show the effectiveness of our proposed method. We will release our code and dataset via github.

\end{abstract}

\section{Introduction}
Sentence summarization, a pivotal task in natural language processing, focuses on creating concise summaries from longer sentences. This area has gained significant attention due to its utility in summarizing events and products, as indicated by various studies. Traditionally, this summarization relied solely on the source sentence. However, with the growing presence of multimedia content combining text and images, recent research has evolved towards multimodal summarization (MMS)~\cite{DBLP:conf/sigir/SongJLZCN22,DBLP:journals/ijautcomp/JingLXYSS23,DBLP:conf/sigir/LinJSLSN23}. This approach integrates visual cues with textual information to enhance the summarization process, making it easier for readers to quickly capture the essence of the information.

In the realm of MMS, notable advancements have been made. \citeauthor{DBLP:conf/ijcai/LiZLZZ18} employed sequence-to-sequence models for better semantic understanding and text generation. They further enhanced MMS by introducing a multimodal selective gate network, which helped in pinpointing the most relevant parts of a sentence based on its corresponding image. More recently, \citeauthor{DBLP:conf/sigir/SongJLZCN22} developed a method for generating efficient product summaries using generative pre-trained language models like BART~\cite{DBLP:conf/acl/LewisLGGMLSZ20}. This method begins by transforming product images into attributes, which are then used by the BART model for generating succinct summaries. This shift towards incorporating visual elements marks a significant development in the field of sentence summarization. A significant challenge faced by these models is the issue of hallucination~\cite{DBLP:conf/emnlp/WanB22}. This occurs when the model produces content that is neither present nor implied in the original input text and image.

Recent progress has been made in creating metrics that align closely with human assessments of factual accuracy in summaries~\cite{DBLP:conf/naacl/TangNWWDWLCMR22, DBLP:conf/naacl/ZhuHXZZHJ21}. These metrics are designed to evaluate the level of factual consistency between the original document and its generated summary. However, there is only one related work that employs factual accuracy in the multimodal summarization task. \citeauthor{DBLP:conf/emnlp/WanB22} proposed CLIPBERTSCORE, a simple weighted combination of CLIPScore~\cite{hessel2021clipscore} and BERTScore~\cite{zhang2020bertscore} to leverage the robustness and strong factuality detection performance between image-summary
and document summary, respectively. 

However, BERTScore pays more attention to more grammatically correct sentences rather than factual sentences~\cite{DBLP:conf/wmt/HannaB21}. Therefore, the factuality of the text input cannot be measured.   In addition, due to CLIPScore's limitations in accurately counting objects~\cite{DBLP:conf/icml/RadfordKHRGASAM21} or conducting compositional reasoning~\cite{DBLP:conf/cvpr/MaHGGGK23}, the CLIPScore often proves to be unreliable and can yield inaccurate results. Furthermore, the existing evaluation metric only generates an overall score, which is coarse-grained and less explainable.

To tackle these limitations, we propose two fine-grained and explainable factuality evaluation frameworks for multimodal summarization (FALLACIOUS), which can be applied to reference-based and reference-free scenarios.  In the reference-based situations, we generate comprehensive questions based on the input textual modality, to provide atomic evaluation. Then, we compare the answers regarding the referenced summary and predicted summary to derive the final factuality score. In the reference-free situation, we generate questions from the model-predicted summary and tested whether the atomic information mentioned in the questions existed in the image and document or not. Based on this, we can also get the final factuality score.

Our contributions can be summarized as: 
(1) We proposed fine-grained and explainable factuality evaluation frameworks for multimodal summarization under reference-based and reference-free scenarios;
(2) We offer an in-depth analysis of our metric along with its components across different benchmarks for evaluating factuality. Through this, we provide substantial empirical proof of its robustness.
\section{Frameworks}
\subsection{Problem Definition}
Suppose we have an image $I$ and a document $D$. The existing multimodal summarization models aim to generate a summary based on the multimodal input (i.e. $I$ and $D$), $S = \mathcal{M}(I, D),$
where $\mathcal{M}$ denotes the multimodal summarization model which takes $I$ and $D$ as inputs and generate a summary. 
Then we devise the evaluation metric to assess the faithfulness of the generated summary $S$. We devised the reference-based and reference-free faithfulness evaluation metrics, respectively.  The former is based on the ground truth summary, formulated as $S_r = \mathcal{F}(I, D, Y, S)$,
where $Y$ is the ground-truth summary. Different from it, the reference-free factuality evaluation metric supposes that there is no available ground truth summary, formulated as $S_s = \mathcal{F}(I, D, S)$. 
In this setting, the metric has more comprehensive application sceneries.
 
\begin{figure}[t]
    \centering
    \includegraphics[width=1\linewidth]{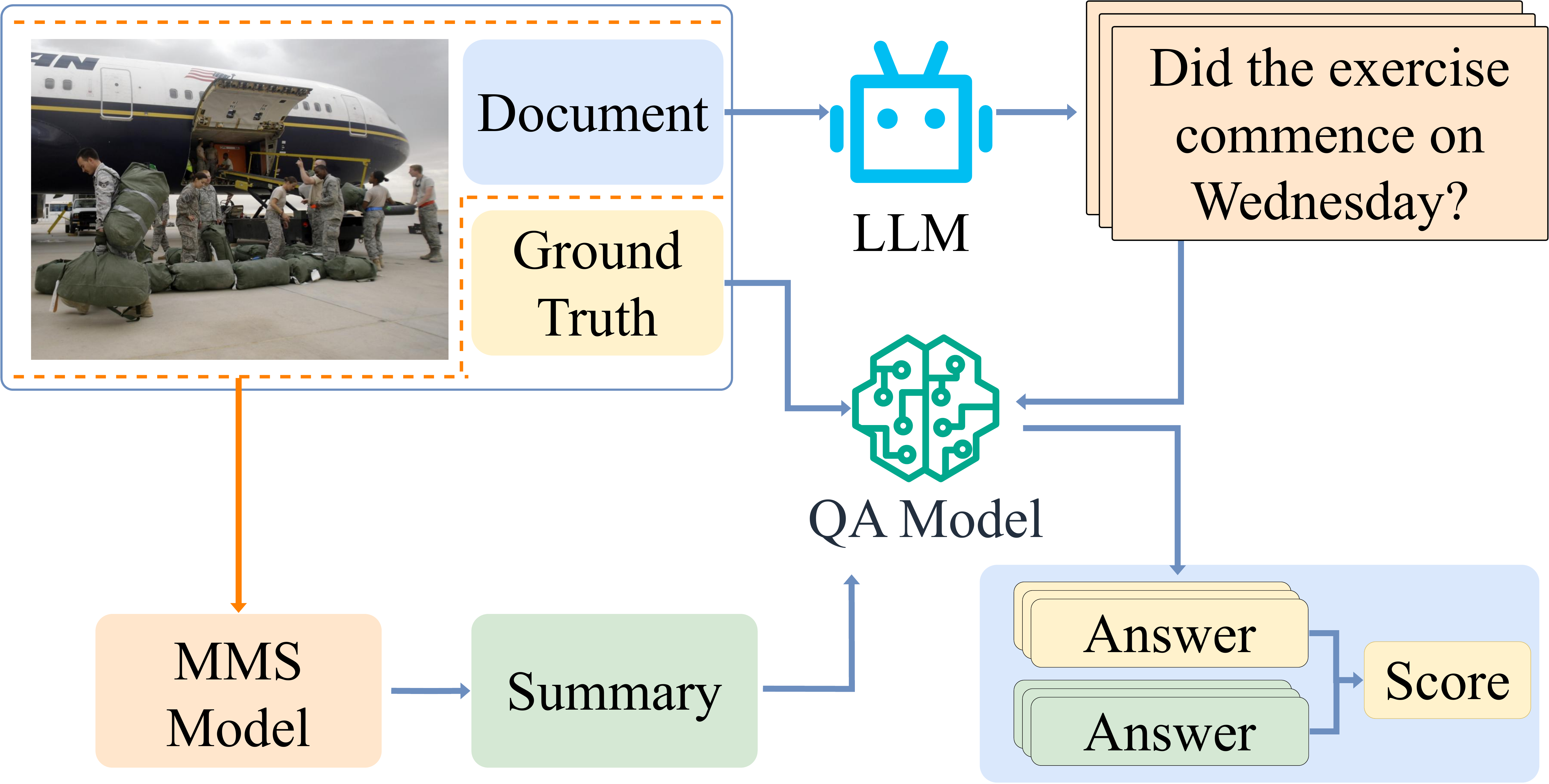}
    \caption{The proposed reference-based framework. 
    }
    \label{fig:with_reference}
\end{figure}

\subsection{Evaluation Framework}
We devise two different evaluation frameworks (i.e. reference-based and reference-free frameworks) to facilitate factuality evaluation under various application scenarios. 

\subsubsection{Reference-based}
As depicted in Figure \ref{fig:with_reference}, the reference-based framework contains three steps: question generation, answer generation, and factuality score aggregation. 

\paragraph{Question Generation} To get a fine-grained evaluation in terms of factuality, we first generate a question set $\mathcal{Q}$ based on the input document, which can generate comprehensive evaluation questions. All the questions are based on atomic information~\cite{DBLP:journals/corr/abs-2305-14251}. The prompt for question generation is detailed in Supplementary Material. 
This process is implemented by a Large Language Model (LLM), formalized as $LLM(Y, D) \rightarrow \mathcal{Q}$.

\paragraph{Answer Generation} Next, we utilize a Question Answering (QA) model to answer the questions $\mathcal{Q}$ based on the ground-truth summary $Y$ and generated summary $S$, respectively. We denote the set of answers as $\mathcal{A}_r$ and $\mathcal{A}_s$ corresponding to the ground-truth summary $Y$ and generated summary $S$. Here, we choose GPT-4 as the model that predicts responses to specific questions, with three possible outcomes: yes, no, or not provided.



\paragraph{Score Aggregation} 
Finally, to evaluate the factuality of the predicted summary $S$, a factuality score $S_{r}$ is calculated. When the answers based on the reference summary and generated summary are consistent, the information that corresponds to the question is truthful. Therefore, this score can be calculated by the consistency between the set of answers $\mathcal{A}_{r}$ and  $\mathcal{A}_{s}$. The score is computed as the sum of individual match scores for all questions, represented by the following equation, 
\begin{equation}
    S_{r} = \frac{\sum_{a_r\in \mathcal{A}_{r}, a_s\in \mathcal{A}_{s}}  \mathbb{I} (a_r==a_s)} {|\mathcal{Q}|},
\end{equation}
where $\mathbb{I}$ is the indicator function and $ |\mathcal{Q}|$ is the number of generated questions.

\begin{figure}[t]
    \centering
    \vspace{-3mm}\includegraphics[width=1\linewidth]{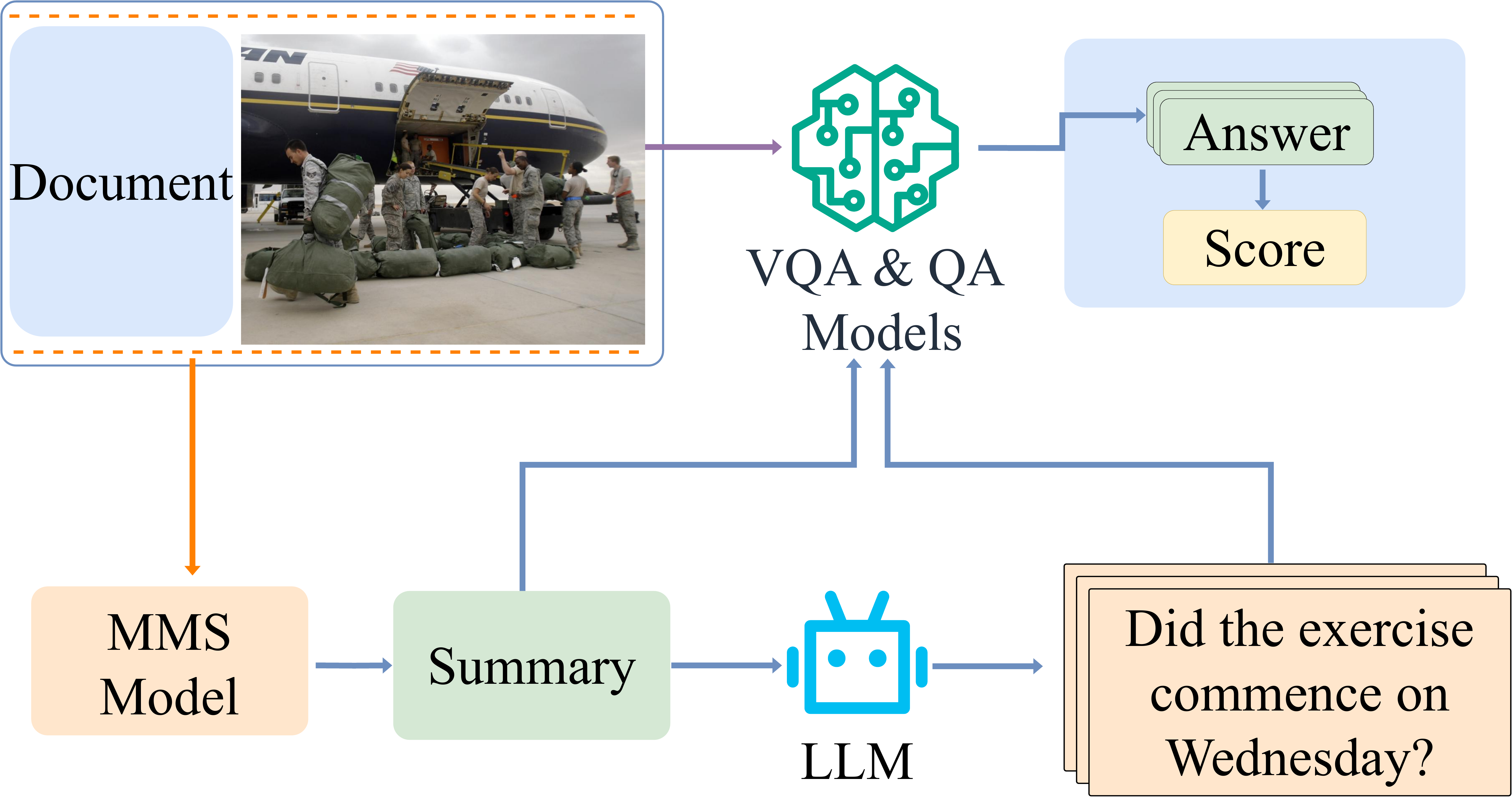}
    \caption{The proposed reference-free framework.
    }
    \label{fig:without_reference}
\end{figure}

\subsubsection{Reference-free}
Obviously, manually annotating ground truth summary for multimodal data is labor- and time-extensive. Hence, we further present a reference-free framework illustrated in Figure \ref{fig:without_reference}, which consists of three stages: question generation, answer generation, and score aggregation. 

\paragraph{Question Generation}
In this situation, the reference summary is not available for evaluation. Therefore, we resort to the consistency between multimodal input and summary to access the model factuality. 
Unlike the reference-based framework, we only generate questions $\mathcal{Q}$ based on the summary $S$ generated by the model.
It ensures that the questions are directly relevant to the content of the summary, allowing for a more precise assessment of the factuality of the predicted summary, even without the reference. 
Similarly, we leverage an LLM to generate questions $\mathcal{Q}$ from the generated summary $S$. This process can be formalized as:
$LLM(S) \rightarrow \mathcal{Q}$

\paragraph{Answer Generation}
To verify the consistency between the multimodal input and summary, we compute the consistency for each modality. Specifically, we use a QA model to answer questions $\mathcal{Q}$ based on the document $D$ and a VQA model to answer questions $\mathcal{Q}$ based on the image $I$. 
The QA model also is GPT-4, and the VQA model is implemented by BLIP-2~\cite{DBLP:conf/icml/0008LSH23}. We denote the set of answers as $\mathcal{A}_t$ and $\mathcal{A}_i$ corresponding to the document $D$ and the image $I$. 

\paragraph{Score Aggregation}
Finally, we aggregate all QA pairs to get the factuality score $S_{s}$, which can be represented as:
\begin{equation}
    S_{s} = \frac{\sum_{a_t \in \mathcal{A}_{t}, a_i \in \mathcal{A}_{i}} \mathbb{I} (H(a_t) \lor H(a_i))}{|\mathcal{Q}|}, 
\end{equation}
where $\mathbb{I}$ denotes the indicator function, $H(x)$ represents a condition where $x$ is equal to `Yes', $\lor$ symbolizes the logical OR operation, and $|\mathcal{Q}|$ is the total number of questions in $\mathcal{Q}$. 

In this formulation, we consider that if either the $a_t$ or the $a_i$ contains `Yes', it indicates factuality, signifying that the model has not fabricated information. Otherwise, it indicates a hallucination.

\section{Experiments}
\begin{table}[]
  \centering
  \caption{Pearson correlation coefficients between automatic metrics and human judgments of factuality on MMSS.}
  \scriptsize
    \begin{tabular}{cccc}
    \toprule
    Metric & Document&Image&Combined \\
    \midrule
        BLEU & 0.15 & - & 0.15 \\
        ROUGE-1 & 0.23 & - & 0.23 \\
        ROUGE-L & 0.13 & - & 0.13 \\
        BERTScore & 0.45 & - & 0.45 \\
    \midrule
        CLIPScore & - & 0.13 & 0.13 \\
    \midrule
        CLIPBERTScore & 0.45 & 0.13 & 0.42 \\ 
    \midrule
     FALLACIOUS (reference-based) & 0.48 &- & \textbf{0.48} \\
     FALLACIOUS (reference-free) & 0.63 & 0.14 & \textbf{0.51} \\
    \bottomrule
    \end{tabular}%
  \label{tab:correlation}%
\end{table}

\begin{table}[]
  \centering
  \caption{Pearson correlation coefficients between automatic metrics and human judgments of factuality on CEPSUM.}
  \scriptsize
    \begin{tabular}{cccc}
    \toprule
    Metric & Document&Image&Combined \\
    \midrule
        BLEU & 0.52 & - & 0.52 \\
        ROUGE-1 & 0.49 & - & 0.49 \\
        ROUGE-L & 0.53 & - & 0.53 \\
        BERTScore & 0.63 & - & 0.63 \\
    \midrule
        CLIPScore & - & 0.21 & 0.21 \\
    \midrule
        CLIPBERTScore & 0.63 & 0.21 & 0.66 \\ 
    \midrule
     FALLACIOUS (reference-based) & 0.78 &- & \textbf{0.78} \\
     FALLACIOUS (reference-free) & 0.78 & 0.31 & \textbf{0.88} \\
    \bottomrule
    \end{tabular}%
  \label{tab:correlation2}%
\end{table}
\subsection{Experimental Setting} 
We employed the BART-MMSS~\cite{DBLP:conf/sigir/LinJSLSN23} model and MMSS dataset~\cite{DBLP:conf/ijcai/LiZLZZ18} to evaluate the effectiveness of our proposed method. 
We employed GPT-4 to generate questions and as the QA model. We used BLIP-2~\cite{BLIP2} as the VQA model. 
\subsection{Human Evaluation}
To verify the effectiveness of the reference-based evaluation framework, we randomly selected 200 samples from the MMSS dataset. 
In addition, we randomly annotated 200 samples generated by another model V2P~\cite{DBLP:conf/sigir/SongJLZCN22} on another dataset CEPSUM~\cite{DBLP:conf/aaai/LiYXWHZ20}. 
For the dataset MMSS, each sample has a maximum of 12 QA pairs, and a minimum of 4 QA pairs, with an average of 8.34 QA pairs per text, totaling 1668 QA pairs. On average, the length of the source sentence
and the target summary are 22 and 8, respectively. 
Similarly, for the reference-free evaluation framework, we also utilized these 200 samples. For this framework, the number of generated questions is 3. 
For the dataset CEPSUM, each sample has a maximum of 15 QA pairs, and a minimum of 5 QA pairs, with an average of 10.2 QA pairs per text, totaling 2040 QA pairs. The average token lengths of the input document and output summary are 316 and 79, respectively.
Similarly, for the reference-free evaluation framework, we also utilized these 200 samples. For this framework, the number of generated questions is 6. 

To explore the reliability of the QA and VQA models, we employ annotators to answer all the generated questions from the two human evaluation datasets. Specifically, the annotator manually annotated answers for each generated question based on the original text and images separately. 
In addition, to verify the effectiveness of our frameworks, for each sample (original text, image, model summary), the annotator subjectively conducted manual scoring for the faithfulness of the model summaries on a scale of 1 to 5, representing completely incorrect, mostly incorrect, roughly equal parts correct and incorrect, mostly correct, and completely correct, respectively.

\subsection{Result Analysis}
\paragraph{Evaluating QA Model} We evaluate the performance of the QA model to get the reliability of this component. We collected all QA pairs from the annotated dataset and got an accuracy of 95.3\%, which shows this component is reliable. 

\paragraph{Evaluating VQA Model} Similarly, to evaluate the performance of the VQA model, we collected all VQA questions from the annotated dataset. The accuracy is 82.7\%, which demonstrates the robustness of this component. 


\begin{figure*}[t]
    \centering
    \begin{subfigure}[t]{0.45\linewidth}
        \centering
        \includegraphics[width=\linewidth]{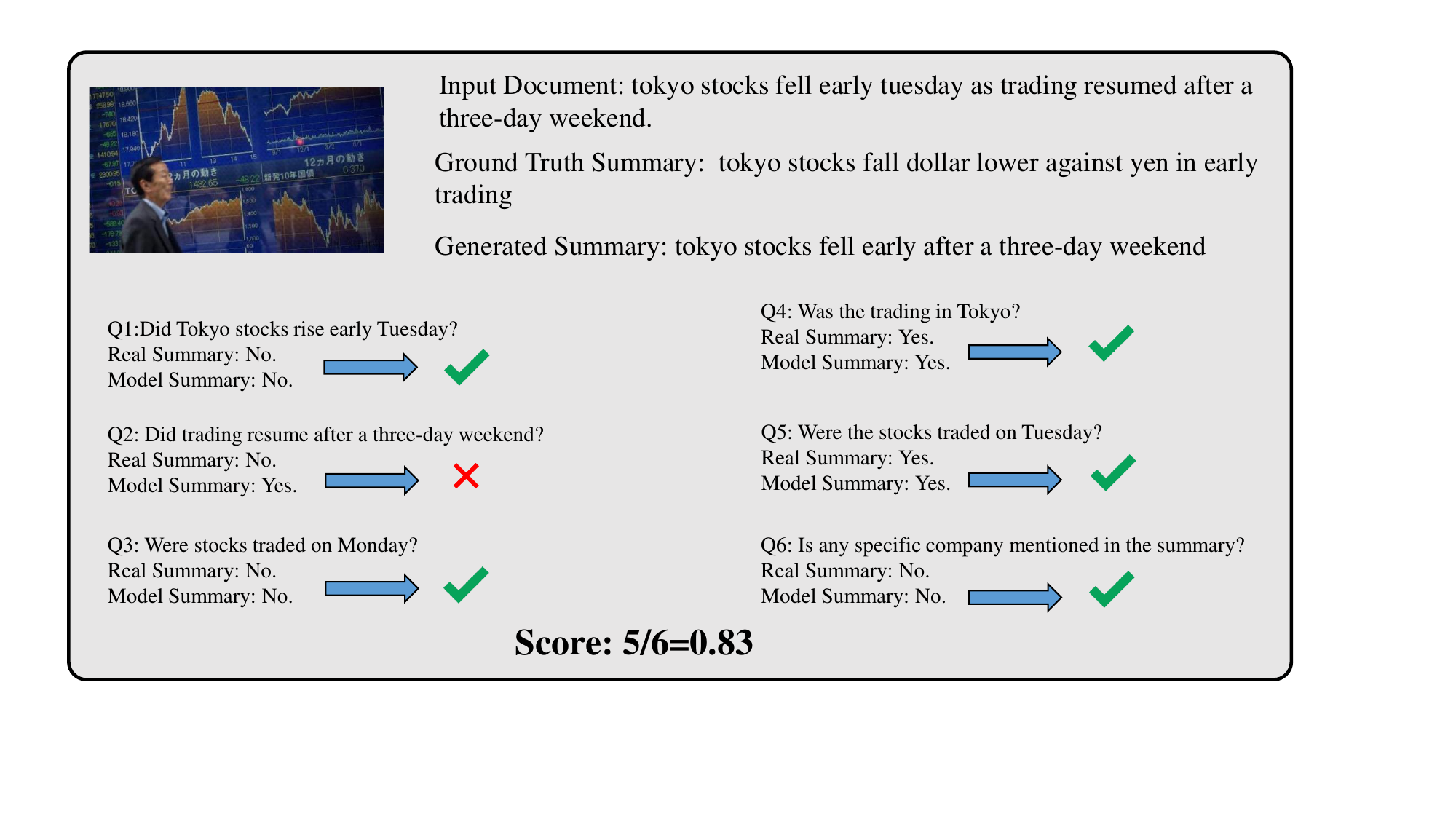}
        \caption{The reference-based evaluation example}
        \label{fig:example3a}
    \end{subfigure}
    \hfill 
    \begin{subfigure}[t]{0.45\linewidth}
        \centering
        \includegraphics[width=\linewidth]{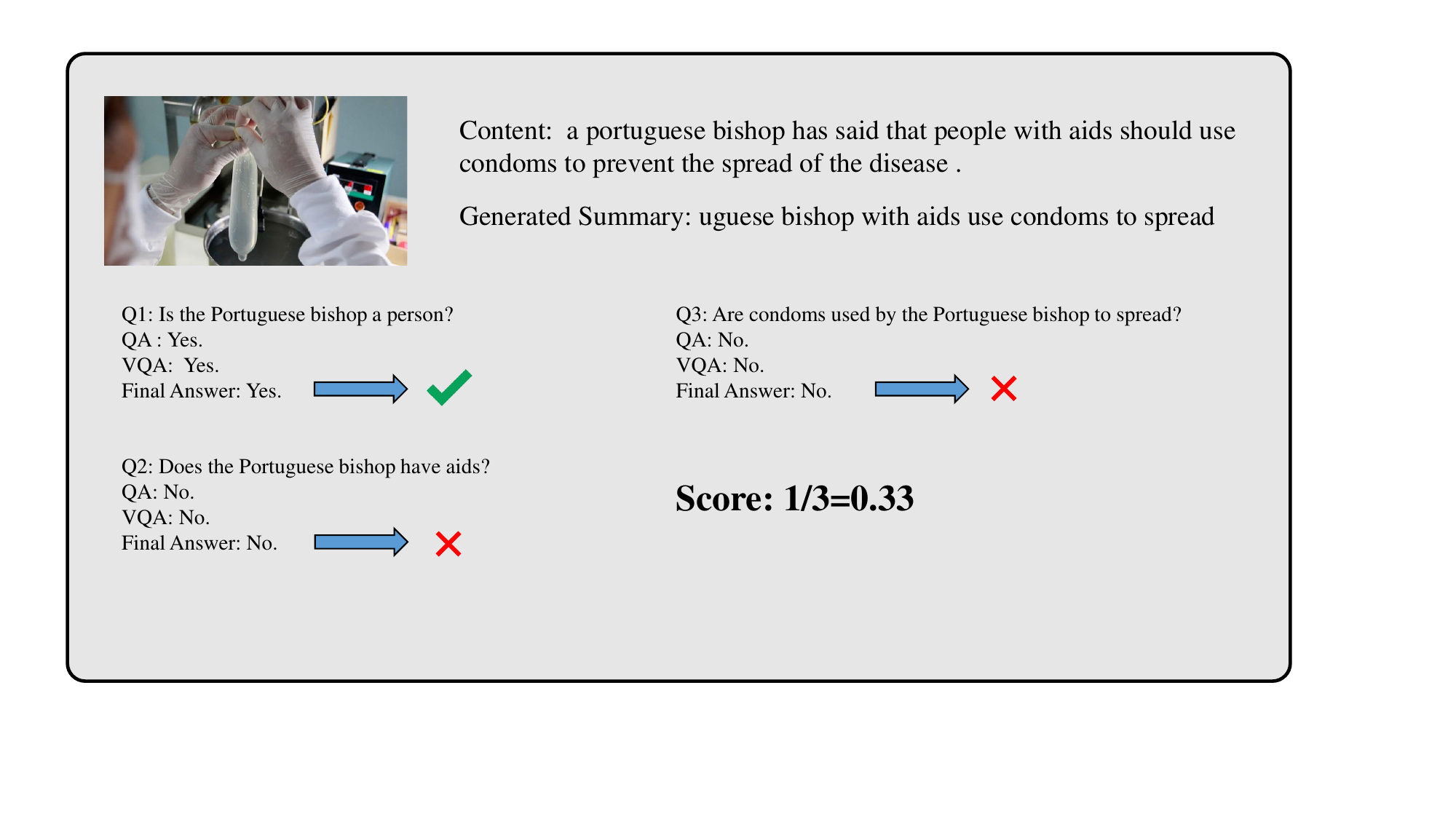}
        \caption{The reference-free evaluation example}
        \label{fig:example3b}
    \end{subfigure}
    \caption{Examples of evaluation methods. QA denotes the answer based on textual modality and VQA denotes the answer based on visual modality. The final answer is the combination of QA and VQA. Score means the score of the proposed metric.}
    \vspace{-5mm}
    \label{fig:example3}
\end{figure*}


\paragraph{Comparison} 
We selected document-based metrics: 1) ROUGE-1~\cite{lin-2004-rouge}, 2) ROUGE-L~\cite{lin-2004-rouge}, 3) BLEU~\cite{papineni2002bleu}, and 4) BertScore~\cite{zhang2020bertscore}; image-based metric: CLIPScore~\cite{hessel2021clipscore};  and combined metric: comCLIPBERTScore~\cite{wan2022evaluating} as baseline metrics. To the best of our knowledge, CLIPBERTScore has been identified as the best metric for evaluating faithfulness in previous assessments of multi-modal summarization. Besides, we also conduct different ablation methods derived from our frameworks based on image input and document input, respectively. 

We computed Pearson correlation coefficients between the human evaluation results and all atomic metrics (all baselines and our proposed metrics) in terms of factuality. 
From Table~\ref{tab:correlation} and \ref{tab:correlation2}, we have two observations: (1) Compared with document-based metrics, image-based metrics always perform worse. The potential reason may be that the summary has more overlapped content with the input text rather than the input image. (2) Our frameworks achieved the best performance compared to other baselines, which demonstrates the superiority of our method.  In addition, we also use the ChatGPT for evaluation on MMSS and achieve a 0.48/0.47 Pearson correlation coefficient for reference-free/based metric which surpasses all the baselines
and shows our framework has the generalization ability. We presented reference-free and reference-based evaluation metrics in Figure \ref{fig:example3a} and  \ref{fig:example3b}.

\paragraph{More Analysis}
For the reference-based framework, we found the ground truth replaces words or rephrases input documents, which makes the generated questions more difficult compared to the reference-free framework. This may cause the performance of a reference-free framework to be better than a reference-based framework.
Compared to text, images show a low correlation with humans, which has also been found in another dataset~\cite{DBLP:conf/emnlp/WanB22}. 
Therefore, we argue that our method can be applied to more datasets. In addition, results on another model V2P, and another dataset CEPSUM show our method is reliable and can be transferred to other datasets. 

CLIPBERTScore is a simple weighted combination of CLIPSCORE and BERTSCore. However, it's important to note that while BERTScore tends to focus more on the grammatical correctness of sentences, it may not necessarily prioritize factual accuracy~\cite{DBLP:conf/wmt/HannaB21}. Furthermore, CLIPScore has its limitations, notably in accurately modeling entities~\cite{DBLP:conf/icml/RadfordKHRGASAM21} and in conducting compositional reasoning~\cite{DBLP:conf/cvpr/MaHGGGK23}. These limitations can lead to reliability issues, often resulting in inaccurate outcomes. As a result, the effectiveness of CLIPBERTScore in delivering accurate factuality scores is potentially compromised. In contrast, our proposed metric offers a more nuanced approach. It is not only fine-grained, allowing for a detailed analysis, but also interpretable, providing clear insights.

\section{Related Work}  

\textbf{Multimodal Summarization} Recently, multimodal summarization, which integrates text with images, has emerged as a significant advancement over traditional text summarization \cite{DBLP:conf/coling/LiZZHZ20,DBLP:conf/aaai/LiYXWHZ20,DBLP:journals/corr/abs-2108-05123,DBLP:conf/emnlp/LiCGCZY20,DBLP:conf/acl/PalaskarLGM19}. Early works focused on combining Convolutional Neural Network-based visual models with Recurrent Neural Network-based textual models to improve the multimodal summarization \cite{DBLP:conf/emnlp/ZhuLL0ZZ18, DBLP:conf/emnlp/ChenZ18, DBLP:conf/ijcai/LiZLZZ18, DBLP:conf/aaai/ZhuZZLZL20, DBLP:conf/aaai/Zhang0P22}. Recent research in the field has shifted towards leveraging pre-trained models for multimodal summarization \cite{DBLP:conf/acl/Jiang00SZ23, DBLP:conf/sigir/LinJSLSN23, DBLP:journals/ijautcomp/JingLXYSS23, DBLP:conf/sigir/SongJLZCN22}. Despite the superior performance of the existing models, these models always suffer from the hallucination where the model generates contents that are not present or entailed by inputs~\cite{DBLP:conf/emnlp/WanB22}. 

\noindent \textbf{Faithfulness and Factuality Metrics} Evaluating the factuality of generated summaries is crucial for their reliability and accuracy. Factuality metrics fall into two main categories: entailment-based metrics and question-answering-based~(QGQA) metrics. Entailment-based metrics~\cite{DBLP:conf/emnlp/KryscinskiMXS20, DBLP:conf/naacl/GoyalD21} assess whether summaries reflect the source document accurately. QGQA approaches \cite{DBLP:conf/acl/DurmusHD20, DBLP:conf/acl/WangCL20, DBLP:conf/emnlp/ScialomDLPSWG21, DBLP:conf/naacl/FabbriWLX22},  evaluate the factuality of the summary by generating and answering input-related questions. Additionally, some other efforts, such as counterfactual estimation \cite{DBLP:conf/emnlp/Xie00LD21} and embedding-based metrics \cite{DBLP:conf/iclr/ZhangKWWA20} have been introduced to this field. Different from them, \citeauthor{DBLP:conf/emnlp/WanB22} is the first one to consider the textual and visual modalities for multimodal summarization factuality evaluation. Although they have achieved compelling success, these methods do not take into account fine-grained elements and are hard to explain. To address this issue, we introduced a framework involving entity-focused questions to evaluate the model's fine-grained factuality, which allows for a more detailed assessment of how accurately the model reflects specific entities and relations within the multimodal input.

\section{Conclusion}
We propose two fine-grained and explainable factuality evaluation frameworks (FALLACIOUS) for multimodal summarization in terms of reference-based and reference-free sceneries.  In the human evaluation, we found that our proposed frameworks can achieve good performance compared with other multimodal summarization metrics. 



\bibliography{aaai25}

\clearpage

\appendix



\definecolor{background}{HTML}{DAE8FC}
\definecolor{user}{HTML}{FFF2CC}
\definecolor{gpt}{HTML}{D5E8D4}

\begin{table*}[]
  \caption{Prompt for question generation and question answering.}
  \label{tab:prompt}
  \setlength{\fboxsep}{5pt} 
  \fbox{ 
    \begin{tabular}{|p{1\textwidth}|}
    \hline
    \rowcolor{background} 
    \textbf{Question Generation for the Task of With Ground Truth:}\\
    \hline
    \rowcolor{user}
    \\
      \rowcolor{user}

      \textbf{User:} Now, you will receive a news article. Please generate questions and provide answers based on the information in the news article. Your questions should be proportional to the amount of information and the number of named entities in the article. The question should be a Yes-No Question.The answers should be "yes," "no," or "not provided," and the answer is based on news article. In other words, you do not need to ask open-ended questions. All questions must have answers, either "yes," "no," or "not provided." Please intentionally ask some incorrect questions to verify the accuracy of the answers. For example, you can intentionally create incorrect answers using names or places that have not appeared before to obtain QA pairs with answers as "NO.". Try to maintain a roughly equal ratio of "yes," "no," and "not provided" answers and cover as much content as possible. The format for providing answers should be [{"Question": question, "Answer": answer}, {"Question": question, "Answer": answer}, {"Question": question, "Answer": answer}, {"Question": question, "Answer": answer}].The news article is:"u.s. , albanian , croatian and macedonian soldiers began a monthlong joint military exercise tuesday , the u.s. embassy said tuesday ." 
     \\
            \rowcolor{user}

      \\

      \hline
      \rowcolor{gpt}
       \\
      \rowcolor{gpt}
      \textbf{GPT:} [
            {
                "Question": "Did the joint military exercise begin on Tuesday?",
                "Answer": "Yes"
            },
            {
                "Question": "Is the joint military exercise only between the U.S. and Albania?",
                "Answer": "No"
            },
            {
                "Question": "Is Croatian military involved in the joint exercise?",
                "Answer": "Yes"
            },
            {
                "Question": "Did the French military participate in the joint exercise?",
                "Answer": "No"
            },
            {
                "Question": "Was the announcement made by the U.S. Embassy?",
                "Answer": "Yes"
            },
            {
                "Question": "Did the joint military exercise last for a year?",
                "Answer": "No"
            },
            {
                "Question": "Did Macedonian soldiers participate in the joint exercise?",
                "Answer": "Yes"
            },
            {
                "Question": "Did the news summary provide information about the purpose of the exercise?",
                "Answer": "No"
            },
            {
                "Question": "Did the exercise commence on Wednesday?",
                "Answer": "No"
            },
            {
                "Question": "Is the joint exercise expected to last for a month?",
                "Answer": "Yes"
            }
        ] \\
        \rowcolor{gpt}
        \\
      \hline
    \rowcolor{background} 
      \textbf{Question Generation for the Task of Without Ground Truth:}\\
      \hline
            \rowcolor{user}
       \\
      \rowcolor{user}
      \textbf{User:} I will now provide you with a news summary. Please generate three yes/no questions regarding the entities and relationships within this sentence. The answers to these three questions should all be true. Please present the questions in the format of ["question1", "question2", "question3"].The summary is:u.s. albanian croatian and macedonian soldiers joint military exercise.s. \\
                  \rowcolor{user}
      \\
      \hline
            \rowcolor{gpt}
             \\
      \rowcolor{gpt}
      \textbf{GPT:} [
            "Are U.S. soldiers participating in a joint military exercise with Albanian soldiers?",
            "Are U.S. soldiers participating in a joint military exercise with Croatian soldiers?",
            "Are U.S. soldiers participating in a joint military exercise with Macedonian soldiers?"
        ] \\
              \rowcolor{gpt}
              \\
      \hline
\rowcolor{background} 
      \textbf{Question Answering:}\\
      \hline
            \rowcolor{user}
             \\
      \rowcolor{user}
      \textbf{User:}I will provide you with a news segment and a question; please provide the answers to the questions in the form of 0 or 1, where 0 represents no and 1 represents yes. news:u.s. , albanian , croatian and macedonian soldiers began a monthlong joint military exercise Tuesday, the u.s. embassy said Tuesday. question: Are U.S. soldiers participating in a joint military exercise with Albanian soldiers?\\
                  \rowcolor{user}
      \\
      \hline
            \rowcolor{gpt}
       \\

      \rowcolor{gpt}
      \textbf{GPT:}1\\
      \rowcolor{gpt}
      \\
      \hline
    \end{tabular}
} 

\end{table*}

\end{document}